# PAC-Bayesian Policy Evaluation
# for Reinforcement Learning

**Mahdi Milani Fard**
School of Computer Science
McGill University
Montreal, Canada
mmilan1@cs.mcgill.ca

**Joelle Pineau**
School of Computer Science
McGill University
Montreal, Canada
jpineau@cs.mcgill.ca

**Csaba Szepesvári**
Department of Computing Science
University of Alberta
Edmonton, Canada
szepesva@ualberta.ca

## Abstract

Bayesian priors offer a compact yet general means of incorporating domain knowledge into many learning tasks. The correctness of the Bayesian analysis and inference, however, largely depends on accuracy and correctness of these priors. PAC-Bayesian methods overcome this problem by providing bounds that hold regardless of the correctness of the prior distribution. This paper introduces the first PAC-Bayesian bound for the batch reinforcement learning problem with function approximation. We show how this bound can be used to perform model-selection in a transfer learning scenario. Our empirical results confirm that PAC-Bayesian policy evaluation is able to leverage prior distributions when they are informative and, unlike standard Bayesian RL approaches, ignore them when they are misleading.

## 1 Introduction

Prior distribution along with Bayesian inference have been used in multiple areas of machine learning to incorporate domain knowledge and impose general variance-reducing constraints, such as sparsity and smoothness, within the learning process. These methods, although elegant and concrete, have often been criticized not only for their computational cost, but also for their strong assumptions on the correctness of the prior distribution. Bayesian guarantees often fail to hold when the inference is performed with priors that are different from the distribution of the underlying true model.

Frequentist methods such as Probably Approximately Correct (PAC) learning, on the other hand, provide distribution-free convergence guarantees (Valiant, 1984). These bounds, however, are often loose and impractical, reflecting the inherent difficulty of the learning problem when no assumptions are made on the distribution of the data.

Both Bayesian and PAC methods have been proposed separately for reinforcement learning (Kearns and Singh, 2002; Brafman and Tennenholtz, 2003; Strehl and Littman, 2005; Kakade, 2003; Duff, 2002; Wang et al., 2005; Poupart et al., 2006; Kolter and Ng, 2009), where an agent is learning to interact with an environment to maximize some objective function. These methods are mostly focused on the so-called exploration–exploitation problem, where one aims to balance the amount of time spent on gathering information about the dynamics of the environment and the time spent acting optimally according to the current estimates. PAC methods are much more conservative and spend more time exploring the system and collecting information. Bayesian methods, on the other hand, are greedier and only solve the problem over a limited planning horizon.

The PAC-Bayesian approach (McAllester, 1999; Shawe-Taylor and Williamson, 1997), takes the best of both worlds by combining the distribution-free correctness of PAC theorems with the data-efficiency of Bayesian inference. PAC-Bayesian bounds do not require the Bayesian assumption to hold. They instead measure the consistency of the prior over the training data, and leverage the prior only when it seems informative. The empirical results of model selection algorithms for classification tasks using these bounds are comparable to some of the most popular learning algorithms, such as AdaBoost and Support Vector Machines (Germain et al., 2009).

Fard and Pineau (2010) introduced the idea of PAC-Bayesian model-selection in reinforcement learning (RL) for finite state spaces. They provided PAC-Bayesian bounds on the approximation error in the value function of stochastic policies when a prior distribution is available either on the space of possible models, or on the space of value functions. Model selection based on these bounds provides a robust use of Bayesian priors outside the Bayesian inference framework. Their work, however, is limited to small and discrete domains, and is mostly useful when sample transitions are drawn uniformly across the state space. This is problematic as most RL domains are relativity large, and require function approximation over continuous state spaces.

This paper provides the first PAC-Bayesian bound for value function approximation on continuous state spaces. We use results by Samson (2000) to handle non i.i.d. data that are collected on Markovian processes, and use this, along with general PAC-Bayesian inequalities, to get a bound on the approximation error of a value function sampled from any distribution over measurable functions. We empirically evaluate these bounds for model selection on two different continuous RL domains in a knowledge-transfer setting. Our results show that a PAC-Bayesian approach in this setting is indeed able to use the prior distribution when it is informative and matches the data, and ignore it when it is misleading.

## 2 Background and Notation

A *Markov Decision Process* (MDP) $M = (\mathcal{X}, \mathcal{A}, T, R)$ is defined by a (possibly infinite) set of states $\mathcal{X}$, a set of actions $A$, a transition probability kernel $T : \mathcal{X} \times \mathcal{A} \to \mathcal{M}(\mathcal{X})$, where $T(.|x,a)$ defines the distribution of next state given that action $a$ is taken in state $x$, and a (possibly stochastic) reward function $R : \mathcal{X} \times \mathcal{A} \to \mathcal{M}([0, R_{\max}])$. Throughout the paper, we focus on finite-action, continuous state, discounted-reward MDPs, with the discount factor denoted by $\gamma \in [0,1)$. At discrete time steps, the reinforcement learning agent chooses an action and receives a reward. The environment then changes to a new state according to the transition kernel.

A *policy* is a (possibly stochastic) function from states to actions. The *value of a state* $x$ for policy $\pi$, denoted by $V^\pi(x)$, is the expected value of the discounted sum of rewards ($\sum_t \gamma^t r_t$) if the agent starts in state $x$ and acts according to policy $\pi$. The value function satisfies the Bellman equation:

$$V^\pi(x) = R(x, \pi(x)) + \gamma \int V^\pi(y) T(dy|x, \pi(x)). \quad (1)$$

There are many methods developed to find the value of a policy (policy evaluation) when the transition and reward functions are known. Among these there are dynamic programming methods in which one iteratively applies the *Bellman operator* (Sutton and Barto, 1998) to an initial guess of the optimal value function. When the transition and reward models are not known, one can use a finite sample set of transitions to learn an approximate value function. Least-squares temporal difference learning (LSTD) and its derivations (Boyan, 2002; Lagoudakis and Parr, 2003) are among the methods used to learn a value function based on a finite sample.

## 3 A general PAC-Bayes bound

We begin by first stating a general PAC-Bayes bound. In the next section, we use this result to derive our main bound for the approximation error in an RL setting.

Let $\mathcal{F}$ be a class of real-valued functions over a well-behaved domain $\mathcal{X}$ (i.e., $\mathcal{X}$ could be a bounded measurable subset of a Euclidean space). For ease of presentation, we assume that $\mathcal{F}$ has countably many functions. For a measure $\rho$ over $\mathcal{F}$, and a functional, $\mathcal{R} : \mathcal{F} \to \mathbb{R}$, we define $\rho \mathcal{R} \stackrel{\text{def}}{=} \int \mathcal{R}(f) d\rho(f)$.

**Theorem 1.** *Let $\mathcal{R}$ be a random functional over $\mathcal{F}$ with a bounded range. Assume that for some $C > 0$, $c > 1$, for any $0 < \delta < 1$ and $f \in \mathcal{F}$, w.p. $1 - \delta$,*

$$\mathcal{R}(f) \leq \sqrt{\frac{\log(C/\delta)}{c}}. \quad (2)$$

*Then, for any measure $\rho_0$ over $\mathcal{F}$, w.p. $1 - \delta$, for all measures $\rho$ over $\mathcal{F}$:*

$$\rho \mathcal{R} \leq \sqrt{\frac{\log(\frac{1+C(c-1)}{\delta}) + K(\rho, \rho_0)}{c-1}}, \quad (3)$$

*where $K(\rho, \rho_0)$ denotes the Kullback-Leibler divergence between $\rho$ and $\rho_0$.*

The proof (included in the appendix) is a straightforward generalization of the proof presented by Boucheron et al. (2005).

## 4 Application to RL

Consider some MDP, with state space $\mathcal{X}$, and a policy $\pi$ whose stationary distribution $\rho^\pi$ exists. Let $\mathcal{D}_n = ((X_i, R_i, X'_{i+1})_{i=1}^n)$ be a random sample of size $n$ such that $X_i \sim \rho^\pi$, $(X'_{i+1}, R_i) \sim P^\pi(\cdot|X_i)$, where $P^\pi$ is the Markov kernel underlying policy $\pi$: The $i^{\text{th}}$ datum $(X_i, R_i, X'_{i+1})$ is an elementary transition from state $X_i$ to state $X'_{i+1}$ while policy $\pi$ is followed and the reward associated with the transition is $R_i$. Further, to simplify the exposition, let $(X, R, X')$ be a transition whose joint distribution is the same as the common joint of $(X_i, R_i, X'_{i+1})$.

Define the functionals $\mathcal{R}, \mathcal{R}_n$ over the space of real-valued, bounded measurable functions over $\mathcal{X}$ as follows: Let $V : \mathcal{X} \to \mathbb{R}$ be such a function. Then

$$\mathcal{R}(V) = \mathbb{E}\left[\left\{R + \gamma V(X') - V(X)\right\}^2\right],$$
$$\mathcal{R}_n(V) = \frac{1}{n} \sum_{i=1}^{n} \left\{R_i + \gamma V(X'_i) - V(X_i)\right\}^2.$$

The functional $\mathcal{R}$ is called the squared sample Bellman error, while $\mathcal{R}_n$ is the empirical squared sample Bellman error. Clearly, $\mathbb{E}[\mathcal{R}_n(V)] = \mathcal{R}(V)$ holds. The following lemma (proved in the appendix) is a concentration bound connecting $\mathcal{R}$ and $\mathcal{R}_n$.

**Lemma 2.** *Under proper mixing conditions for the sample, and assuming that the random rewards are sub-Gaussian, there exists constants $c_1 > 0$, $c_2 \geq 1$ which depend only on $P^\pi$ such that for any $V_{\max} > 0$, for any measurable function $V$ bounded by $V_{\max}$, and any $0 < \delta < 1$, w.p. $1 - \delta$,*

$$\mathcal{R}(V) - \mathcal{R}_n(V) \leq \sqrt{\frac{V_{\max}^2 c_1}{n} \log\left(\frac{c_2}{\delta}\right)}. \quad (4)$$

Hence, by Theorem 1, for any countable set $\mathcal{F}$ of functions $V$ bounded by $V_{\max}$, for any distribution $\mu_0$ over these functions, if $n > V_{\max}^2 c_1$, then for all $0 < \delta < 1$, w.p. $1 - \delta$, for all measures $\mu$ over $\mathcal{F}$:

$$\mu(\mathcal{R} - \mathcal{R}_n) \leq \sqrt{\frac{\log\left(\frac{c_2 n}{c_1 V_{\max}^2 \delta}\right) + K(\mu, \mu_0)}{\frac{n}{V_{\max}^2 c_1} - 1}}. \quad (5)$$

Now, we show how this bound can be used to derive a PAC-Bayes bound on the error of a value function $V$ that is drawn from an arbitrary distribution over measurable functions. For a distribution $\rho$ over the state space $\mathcal{X}$, let $\|\cdot\|_\rho$ be the weighted $L^2$ norm: $\|V\|_\rho^2 = \int [V(x)]^2 d\rho(x)$. Further, let $B^\pi$ be the Bellman operator underlying $\pi$: $B^\pi V(x) = \mathbb{E}[R + \gamma V(X')|X = x]$. Fix some $V$. Since $B^\pi$ is a $\gamma$-contraction w.r.t. the norm $\|\cdot\|_{\rho^\pi}$, a standard argument shows that (Bertsekas and Tsitsiklis, 1996):

$$\|V - V^\pi\|_{\rho^\pi} \leq \frac{\|B^\pi V - V\|_{\rho^\pi}}{1 - \gamma}. \quad (6)$$

Now, using the variance decomposition $\mathbb{V}\mathrm{ar}[U] = \mathbb{E}[U^2] - \mathbb{E}[U]^2$, we get $\mathcal{R}(V) = \|B^\pi V - V\|_{\rho^\pi}^2 + \Gamma_\pi(V)$, where $\Gamma_\pi(V) = \mathbb{E}[\mathbb{V}\mathrm{ar}[R + \gamma V(X')|X]]$ (see, e.g., (Antos et al., 2008))[1]. Thus, for $\varepsilon_\pi^2(V) = \|V - V^\pi\|_{\rho^\pi}^2$,

$$\varepsilon_\pi^2(V) \leq \frac{1}{(1-\gamma)^2} [\mathcal{R}(V) - \Gamma_\pi(V)]. \quad (7)$$

Combining this inequality with (5), we get the following result:

**Theorem 3.** *Fix a countable set $\mathcal{F}$ of real-valued, measurable functions with domain $\mathcal{X}$, which are bounded by $V_{\max}$. Assume that the conditions of Lemma 2 hold and let $c_1, c_2$ be as in this lemma. Fix any measure $\mu_0$ over these functions. Assume that $n > V_{\max}^2 c_1$. Then, for all $0 < \delta < 1$, with probability $1 - \delta$, for all measures $\mu$ over $\mathcal{F}$:*

$$\mu \varepsilon_\pi^2 \leq \frac{1}{(1-\gamma)^2} \left\{ \mu \mathcal{R}_n + \sqrt{\frac{\log\left(\frac{c_2 n}{c_1 V_{\max}^2 \delta}\right) + K(\mu, \mu_0)}{\frac{n}{V_{\max}^2 c_1} - 1}} - \mu \mathbb{E}[\Gamma_\pi] \right\}.$$

*Further, the same bound holds for $\|\bar{V}_\mu - V^\pi\|_{\rho^\pi}^2$, where $\bar{V}_\mu = \int V d\mu(V)$ is the $\mu$-average of value functions from $\mathcal{F}$.*

*Proof.* The first statement follows from (7) combined with (5), as noted earlier. To see this just replace $\mathcal{R}(V)$ in (7) with $\mathcal{R}(V) - \mathcal{R}_n(V) + \mathcal{R}_n(V)$. Then, integrate both sides with respect to $\mu$ and apply (5) to bound $\mu(\mathcal{R} - \mathcal{R}_n)$. The second part follows from the first part, Fubini's theorem and Jensen's inequality: $\|\bar{V}_\mu - V^\pi\|_{\rho^\pi}^2 = \|\int (V^\pi - V) d\mu(V)\|_{\rho^\pi}^2 \leq \int \|V^\pi - V\|_{\rho^\pi}^2 d\mu(V) = \mu \varepsilon_\pi^2$. □

The theorem bounds the expected error of approximating $V^\pi$ with a value function drawn randomly from some distribution $\mu$. Note that in this theorem, $\mu_0$ must be a fixed distribution, chosen *a priori* (i.e. prior distribution), *but $\mu$ can be chosen in a data dependent manner*, i.e., it can be a "posterior" distribution.

Notice that there are three elements to the above bound (right hand side). The first term is the empirical component of the bound, which enforces the selection of solutions with smaller empirical Bellman residuals. The second term is the Bayesian component of the bound, which penalizes distributions that are far from the prior. The third term corrects for the variance in the return at each state.

---
[1] Here, $\mathbb{V}\mathrm{ar}[U|V] = \mathbb{E}[(U - \mathbb{E}[U|V])^2|V]$ is the conditional variance of $U$ as usual. We shall also use the similarly defined conditional covariance, $\mathbb{C}\mathrm{ov}[(U_1, U_2)|V] = \mathbb{E}[(U_1 - \mathbb{E}[U_1|V])(U_2 - \mathbb{E}[U_2|V])^\top|V]$. When $U_1 = U_2$, we will also use $\mathbb{C}\mathrm{ov}[U_1|V] \stackrel{\text{def}}{=} \mathbb{C}\mathrm{ov}[(U_1, U_2)|V]$.

If we can empirically estimate the right hand side of the above inequality, then we can use the bound in an algorithm. For example, we can derive a PAC-Bayesian model-selection algorithm that searches in the space of posteriors $\mu$ so as to minimize the upper bound.

### 4.1 Linearly parametrized classes of functions

Theorem 3 is presented for any countable families of functions. One can extend this result to sufficiently regular classes of functions (which can carry measures) without any problems.[2] Here we consider the case where $\mathcal{F}$ is the class of linearly parametrized functions with bounded parameters,

$$\mathcal{F}_C = \{ \theta^\top \phi : \|\theta\| \leq C \}$$

where $\phi : \mathcal{X} \to \mathbb{R}^d$ is some measurable function such that $F_{\max} \stackrel{\text{def}}{=} \sup_{x \in \mathcal{X}} \|\phi(x)\|_2 < \infty$. In this case, the measures can be put on the ball $\{ \theta : \|\theta\| \leq C \}$.

Let us now turn to the estimation of the variance term. Assuming that the reward for each transition is independent of the next state, one gets

$$\mathbb{V}\text{ar}\left[R + \gamma V(X')|X\right] = \mathbb{V}\text{ar}\left[R|X\right] + \gamma^2 \mathbb{V}\text{ar}\left[V(X')|X\right].$$

Now, if $V_\theta = \phi^\top \theta$, then:

$$\mathbb{V}\text{ar}\left[V(X')|X\right] = \theta^\top \mathbb{C}\text{ov}\left[\phi(X')|X\right] \theta.$$

Assuming homoscedastic variance for the rewards, and defining $\sigma_R^2 = \mathbb{V}\text{ar}[R]$ and $\Sigma_\phi = \mathbb{E}\left[\mathbb{C}\text{ov}\left[\phi(X')|X\right]\right]$, we get:

$$\mathbb{V}\text{ar}\left[R + \gamma V(X')|X\right] = \sigma_R^2 + \gamma^2 \theta^\top \Sigma_\phi \theta.$$

### 4.2 Estimating the constants

In some cases the terms $\sigma_R^2$ and $\Sigma_\phi$ are known (e.g., $\sigma_R^2 = 0$ when the rewards are a deterministic function of the start state and action, and $\Sigma_\phi = 0$ when the dynamics is deterministic). An alternative is to estimate these terms empirically. This can be done by, e.g., double sampling of next states (assuming one has access to a generative model, or if one can reset the state).[3] If such estimates are generated based on finite sample sets, then we might need to add extra deviation terms to the bound of Theorem 3. For simplicity, we assume that these terms are either known or can be estimated on a separate dataset of many transitions. Examples of such cases are studied in the empirical results.

The constant $c_2$, which depends on the mixing condition of the process, can also be estimated if we have access to a generative model. There are upper bounds for $c_2$ when the sample is a collection of independent trajectories of length less than $h$.

## 5 Empirical Results

In this section, we investigate how the bound of Theorem 3 can be used in a model selection mechanism *for transfer learning* in the RL setting. One experiment is presented on the well-known mountain car problem, the other focuses on a generative model of epileptic seizures built from real-world data.

### 5.1 Case Study: Mountain Car

We design a transfer learning experiment on the Mountain Car domain (Sutton and Barto, 1998), where the goal is to drive an underpowered car beyond a certain altitude up a mountain. We refer the reader to the reference for details of the domain. We learn the optimal policy (name it $\pi$) on the original Mountain Car problem ($\gamma = 0.9$, $reward = 1$ passed the goal threshold and 0 otherwise). Note that the reward and the dynamics are deterministic, therefore $\sigma_R^2 = 0$ and $\Sigma_\phi = 0$. The task is to learn the value function on the original domain, and use that knowledge in similar (though not identical) environments to accelerate the learning process in those new environments. (The other environments will be described later.)

We estimate the value of $\pi$ on the original domain with tile coding (4 tiles of size $8 \times 8$). Let $\theta_0$ be the LSTD solution on a very large sample set in the original domain. To transfer the domain knowledge from this problem, we construct a prior distribution $\mu_0$: product of Gaussians with mean $\theta_0$ and variance $\sigma_0^2 = 0.01$.

In a new environment, we collect a set of trajectories (100 trajectories of length 5), and search in the space of $\lambda$-parametrized posterior measures, defined as follows: measure $\mu_\lambda$ is the product of Gaussians with mean $\left(\frac{\lambda \theta_0}{\sigma_0^2} + \frac{\hat\theta}{\hat\sigma^2}\right) \Big/ \left(\frac{\lambda}{\sigma_0^2} + \frac{1}{\hat\sigma^2}\right)$ and variance $\left(\frac{\lambda}{\sigma_0^2} + \frac{1}{\hat\sigma^2}\right)^{-1}$, where $\hat\theta$ is the LSTD solution based on the sample set on the new environment, and $\hat\sigma^2$ (variance of the empirical estimate) is set to 0.01. The search for the best $\lambda$-parameterized posterior is driven by our proposed PAC-Bayes upper bound on the approximation error. When $\lambda = 0$, $\mu_\lambda$ will be a purely empirical estimate, whereas when $\lambda = 1$, we get the Bayesian posterior for the mean of a Gaussian with known variance (standard Bayesian inference with empirical priors).

---

[2] The extension presents only technical challenges, but leaves the result intact and hence is omitted.

[3] Alternately, one can co-estimate the mean and variance terms (Sutton et al., 2009), keeping a current guess of them and updating both estimates as new transitions are observed.

Note that because the Mountain Car is a deterministic domain, the variance term of Theorem 3 is 0. As we use trajectories with known length, we can also bound the other constants in the bound and evaluate the bound completely empirically based on the observed sample set.

We test this model-selection method on two new environments. The first is a mountain domain very similar to the original problem, where we double the effect of the acceleration of the car. The true value function of this domain is close the original domain, and so we expect the prior to be informative (and thus $\lambda$ to be close to 1). In the second domain, we change the reward function such that it decreases, inversely proportional to the car's altitude: $r(x) = 1 - h(x)$, where $h(x) \in [0, 1]$ is the normalized altitude at state $x$. The value function of $\pi$ under this reward function is largely different from that of the original one, which means that the prior distribution is misleading, and the empirical estimate should be more reliable (and $\lambda$ close to 0).

Table 1 reports the average true error of approximating $V^\pi$ using different methods over 100 runs (purely empirical method is when $\lambda = 0$, Bayesian is when $\lambda = 1$). This corresponds to the left hand side of Theorem 3 for these methods. For the similar environment, the PAC-Bayes bound is minimized consistently with $\lambda = 1$, indicating that the method is fully using the Bayesian prior. The error is thus decreased to less than a half of that of the empirical estimate. For the environment with largely different reward function, standard Bayesian inference results in poor approximation, whereas the PAC-Bayes method is selecting small values of $\lambda$ and is mostly ignoring the prior.

Table 1: Error in the estimated value function $V^\pi$ ($\int \|V - V^\pi\|^2_{\rho^\pi} d\mu(V)$) on the Mountain Car domain. The last row shows the value of the $\lambda$ parameter selected by the PAC-Bayesian method.

|  | Similar Env | Different Env |
| --- | --- | --- |
| Purely empirical | $2.35 \pm 0.12$ | $0.03 \pm 0.01$ |
| Bayesian | $1.03 \pm 0.09$ | $2.38 \pm 0.05$ |
| PAC-Bayes | $1.03 \pm 0.09$ | $0.07 \pm 0.01$ |
| $\lambda_{\text{PAC-Bayes}}$ | 1 | $0.06 \pm 0.01$ |

To further investigate how the value function estimate changes with these different methods, we consider an estimate of the value for the state when the car is at the bottom of the hill. This point estimate is constructed from the PAC-Bayes estimate using the value function obtained by using only the mean of $\mu_\lambda$. To get a sense of the dependence of this estimate on the randomness of the sample the estimate is constructed over 100 runs. We also obtain these estimates using a Bayes estimate and purely empirical estimate. Figure 5.1(left) shows a normal fit to the histogram of the resulting estimates, for the purely empirical and PAC-Bayes estimates. As it can be seen, the distribution of PAC-Bayes estimates (which coincides with the Bayesian posterior as the best $\lambda$ is consistently 1 in this case) is centered around the correct value, but is more peaked than the empirical distribution. This shows that the method is using the prior to converge faster to the correct value.

Figure 5.1(right) compares the distribution of the estimated values for the highly different environment. We can see that, as expected, the Bayesian estimate is heavily biased due to the use of a misleading prior. The PAC-Bayes estimate is only slightly biased away from the empirical one with the same variance on the value. Again, this confirms that PAC-Bayes model-selection is largely ignoring the prior when the prior is misleading.

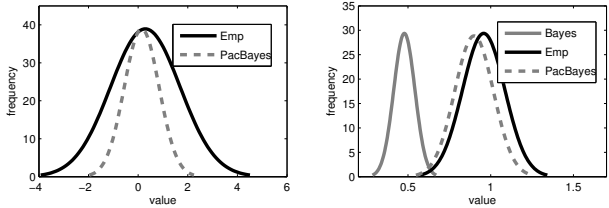

Figure 1: Distribution of the estimated value function on similar (left) and different (right) environments

## 5.2 Case Study: Epilepsy Domain

We also evaluate our method on a more complex domain. The goal of the RL agent here is to apply direct electrical neurostimulation such as to suppress epileptiform behavior. We use a generative model constructed from real-world data collected on slices of rat brain tissues (Bush et al., 2009); the model is available in the RL-Glue framework. Observations are generated over a 4-dimensional real-valued state space. The action choice corresponds to selecting the frequency at which neurostimulation is applied. The reward is $-1$ for steps when a seizure is occurring, $-1/40$ for each stimulation pulse, and 0 otherwise.

We first apply the best clinical fixed rate policy (stimulation is applied at a consistent 1Hz) to collect a large sample set (Bush et al., 2009). We then use LSTD to learn a linear value function over the original feature space. Similar to the experiment described above, we construct a prior (with a similar mean and variance structure), and use it for knowledge transfer in two new cases. This time, we keep the dynamics and reward function intact and instead change the policy. The first modified policy we consider applies stimulation at a fixed rate of 2Hz; this is expected to have a similar value function as the original (1Hz) policy. The other policy we consider applies no stimulation; this is expected to have a very different value function as the seizures are not suppressed.

Table 2: The error of value-function estimates ($\int \|V - V^\pi\|^2_{\rho^\pi} d\mu(V)$) on the Epilepsy domain.

|  | 2 Hz Stimulation | No Stimulation |
|---|---|---|
| Empirical | $0.0044 \pm 0.0007$ | $0.54 \pm 0.06$ |
| Bayesian | $0.0013 \pm 0.0001$ | $0.86 \pm 0.07$ |
| PAC-Bayes | $0.0022 \pm 0.0004$ | $0.69 \pm 0.08$ |
| $\lambda_{\text{PAC-Bayes}}$ | $0.62 \pm 0.05$ | $0.30 \pm 0.05$ |

We sample 10,000 on-policy trajectories of length 1 and use them with the PAC-Bayes model-selection mechanism described previously (with similar $\lambda$-parametrized posterior family on the $\theta$ parameters, $\gamma = 0.8$) to get estimates of the value function. Table 2 summarizes the performance of different methods on the evaluation of the new policies (averaged over 50 runs). The results are not as polarized as those of the Mountain Car experiment, partly because the domain is noisier, and because the prior is neither exclusively informative or misleading. Nonetheless, we observe that the PAC-Bayes method is using the prior more ($\lambda$ averaging around 0.62) in the case of the 2Hz policy, which is consistent with clinical evidence showing that 1Hz and 2Hz have similar effect (Bush et al., 2009), whereas the prior is considered less ($\lambda$ averaging around 0.30) in the case of the 0Hz policy which has substantially (though not entirely) different effects.

## 6 Discussion

This paper introduces the first PAC-Bayesian bound for policy evaluation with function approximation and general state spaces. We demonstrate how such bounds can be used for value function estimation based on finite sample sets. Our empirical results show that PAC-Bayesian model-selection uses prior distributions when they are informative, and ignores them when they are misleading. Our results thus far focus on the policy evaluation case. This approach can be used in a number of applications, including transfer learning, as explored above.

Model-selection based on error bounds has been studied previously with regularization techniques (Farahmand et al., 2009). These bounds are generally tighter for point estimates, as compared to the distributions used in this work. However, our method is more general as it could incorporate arbitrary domain knowledge into the learning algorithm with any type of prior distribution. It can of course use sparsity or smoothness priors, which correspond to well-known regularization methods.

An alternative is to derive margin bounds, similar to those of large-margin classifiers, using PAC-Bayes techniques. This was recently done by Fard and Pineau (2010) in the discrete case. The extension to continuous domains with general function approximation is an interesting future work.

This work does not address the application of PAC-Bayes bounds to derive exploration strategies for the RL problem. Seldin et al. (2011b,a) have studied the exploration problem for multiarmed bandits and have provided algorithms based on PAC-Bayes analysis of martingales. Extensions to contextual bandits and more general RL settings remain interesting open problems.

## Acknowledgements

This work was supported in part by AICML, AITF (formerly iCore and AIF), the PASCAL2 Network of Excellence under EC (grant no. 216886), the NSERC Discovery Grant program and the National Institutes of Health (grant R21 DA019800).

## References

A. Antos, Cs. Szepesvári, and R. Munos. Learning near-optimal policies with Bellman-residual minimization based fitted policy iteration and a single sample path. *Machine Learning*, 71(1):89–129, April 2008.

D. P. Bertsekas and J. N. Tsitsiklis. *Neuro-Dynamic Programming (Optimization and Neural Computation Series, 3)*. Athena Scientific, 1996. ISBN 1886529108.

S. Boucheron, O. Bousquet, and G. Lugosi. Theory of classification: A survey of some recent advances. *ESAIM: Probability and Statistics*, 9:323–375, 2005.

J. A. Boyan. Technical update: Least-squares temporal difference learning. *Machine Learning*, 49(2):233–246, 2002.

R. I. Brafman and M. Tennenholtz. R-max – A general polynomial time algorithm for near-optimal reinforcement learning. *The Journal of Machine Learning Research*, 3:213–231, 2003.

K. Bush, J. Pineau, A. Guez, B. Vincent, G. Panuccio, and M. Avoli. Dynamic Representations for Adaptive Neurostimulation Treatment of Epilepsy. *4th International Workshop on Seizure Prediction*, 2009.

M. O. G. Duff. *Optimal learning: Computational procedures for Bayes-adaptive Markov decision processes*. PhD thesis, University of Massachusetts Amherst, 2002.

A. M. Farahmand, M. Ghavamzadeh, Cs. Szepesvári, and S. Mannor. Regularized policy iteration. In *NIPS*, 2009.

M. M. Fard and J. Pineau. PAC-Bayesian model selection for reinforcement learning. In *NIPS*, 2010.

P. Germain, A. Lacasse, F. Laviolette, and M. Marchand. PAC-Bayesian learning of linear classifiers. In *ICML*, 2009.

S. M. Kakade. *On the sample complexity of reinforcement learning*. PhD thesis, University College London, 2003.

M. Kearns and S. Singh. Near-optimal reinforcement learning in polynomial time. *Machine Learning*, 49(2-3):209–232, 2002.

J. Z. Kolter and A. Y. Ng. Near-Bayesian exploration in polynomial time. In *ICML*, pages 513–520, 2009.

M. G. Lagoudakis and R. Parr. Least-squares policy iteration. *Journal of Machine Learning Research*, 4:1107–1149, 2003. ISSN 1532-4435.

D. A. McAllester. Some PAC-Bayesian theorems. *Machine Learning*, 37(3):355–363, 1999.

S. Meyn and R.L. Tweedie. *Markov Chains and Stochastic Stability*. Cambridge University Press, New York, NY, USA, 2009.

P. Poupart, N. Vlassis, J. Hoey, and K. Regan. An analytic solution to discrete bayesian reinforcement learning. In *ICML*, 2006.

P. M. Samson. Concentration of measure inequalities for Markov chains and $\phi$-mixing processes. *Annals of Probability*, 28(1):416–461, 2000.

Y. Seldin, N. Cesa-Bianchi, F. Laviolette, P. Auer, J. Shawe-Taylor, and J. Peters. PAC-Bayesian analysis of the exploration-exploitation trade-off. In *On-line Trading of Exploration and Exploitation 2, ICML-2011 workshop*, 2011a.

Y. Seldin, F. Laviolette, J. Shawe-Taylor, J. Peters, and P. Auer. PAC-Bayesian analysis of martingales and multiarmed bandits. Technical report, http://arxiv.org/abs/1105.2416, 2011b.

J. Shawe-Taylor and R. C. Williamson. A PAC analysis of a Bayesian estimator. In *COLT*, 1997.

A. L. Strehl and M. L. Littman. A theoretical analysis of model-based interval estimation. In *ICML*, 2005.

R. S. Sutton and A. G. Barto. *Reinforcement Learning: An Introduction*. MIT Press, Cambridge, MA, 1998.

R. S. Sutton, H. R. Maei, D. Precup, S. Bhatnagar, D. Silver, Cs. Szepesvári, and E. Wiewiora. Fast gradient-descent methods for temporal-difference learning with linear function approximation. In *ICML*, 2009.

L. G. Valiant. A theory of the learnable. *Commun. ACM*, 27(11):1134–1142, 1984.

T. Wang, D. Lizotte, M. Bowling, and D. Schuurmans. Bayesian sparse sampling for on-line reward optimization. In *ICML*, 2005.

# 7 Appendix

## 7.1 Proof of Theorem 1

*Proof.* Let $\Delta$ be some functional over $\mathcal{F}$. For some real-valued function over the reals $g$, we will use $g(\Delta)$ to denote the functional over $\mathcal{F}$ which maps $f \in \mathcal{F}$ to $g(\Delta(f))$. When $g$ is measurable, this allows us to write $\rho_0 g(\Delta)$, where $\rho_0$ is a measure over $\mathcal{F}$.

For any functional $\Delta$ over $\mathcal{F}$, by the convex duality of relative entropy, we have

$$\rho \Delta \leq \inf_{\lambda > 0} \frac{1}{\lambda} \left\{ \log\left(\rho_0 e^{\lambda \Delta}\right) + K(\rho, \rho_0) \right\}.$$

We can apply this to $\Delta(f) = (\mathcal{R}(f))_+^2$. Let $\lambda$ be a positive real to be chosen later. Then, we will see that it will be sufficient to bound the right tail probabilities of $\rho_0 e^{\lambda \Delta(f)}$. By Markov's inequality and Fubini, for $\varepsilon > 0$,

$$\mathbb{P}\left(\rho_0 e^{\lambda \Delta} \geq \varepsilon\right) \leq \varepsilon^{-1} \mathbb{E}\left[\rho_0 e^{\lambda \Delta}\right] = \varepsilon^{-1} \rho_0 \mathbb{E}\left[e^{\lambda \Delta}\right],$$

where we used the boundedness of $\mathcal{R}$. Now, fix some $f \in \mathcal{F}$. Then,

$$\begin{aligned}
\mathbb{E}\left[e^{\lambda \Delta(f)}\right] &= 1 + \int_1^\infty \mathbb{P}\left(e^{\lambda \Delta(f)} \geq t\right) dt \\
&= 1 + \int_0^\infty \mathbb{P}\left(\lambda \Delta(f) \geq t\right) e^t dt \\
&= 1 + \int_0^\infty \mathbb{P}\left(\mathcal{R}(f) \geq \sqrt{\frac{t}{\lambda}}\right) e^t dt \\
&\leq 1 + C \int_0^\infty e^{-c(t/\lambda)+t} dt \qquad \text{by (2)} \\
&= 1 + \frac{C}{c/\lambda - 1} = 1 + C(c - 1),
\end{aligned}$$

where in the last step we have chosen $\lambda = c - 1$. Hence, $\mathbb{P}\left(\rho_0 e^{\lambda \Delta} \geq \varepsilon\right) \leq \frac{1 + C(c-1)}{\varepsilon}$. Let $\varepsilon = (1 + C(c-1))/\delta$ to get $\mathbb{P}\left(\log \rho_0 e^{\lambda \Delta} \geq \log((1 + C(c-1))/\delta)\right) \leq \delta$. Hence,

$$\rho \Delta \leq \frac{\log((1 + C(c-1))/\delta) + K(\rho, \rho_0)}{c - 1}$$

holds w.p. $1 - \delta$. The proof is finished by noting that

$$\rho(\mathcal{R})_+ \leq \sqrt{\rho(\mathcal{R})_+^2} = \sqrt{\rho \Delta}.$$

$\square$

## 7.2 Proof of Lemma 2

In this section we give an extension of Bernstein's inequality based on Samson (2000).

Let $X_1, \ldots, X_n$ be a time-homogeneous Markov chain with transition kernel $P(\cdot|\cdot)$ taking values in some measurable space $\mathcal{X}$. We shall consider the concentration of the average of the Hidden-Markov Process

$$(X_1, f(X_1)), \ldots, (X_n, f(X_n)),$$

where $f : \mathcal{X} \to [0, B]$ is a fixed measurable function. To arrive at such an inequality, we need a characterization of how fast $(X_i)$ forgets its past.

For $i > 0$, let $P^i(\cdot|x)$ be the $i$-step transition probability kernel: $P^i(A|x) = \mathbb{P}(X_{i+1} \in A \mid X_1 = x)$ (for all $A \subset \mathcal{X}$ measurable). Define the upper-triangular matrix $\Gamma_n = (\gamma_{ij}) \in \mathbb{R}^{n \times n}$ as follows:

$$\gamma_{ij}^2 = \sup_{(x,y) \in \mathcal{X}^2} \left\| P^{j-i}(\cdot|x) - P^{j-i}(\cdot|y) \right\|_{\text{TV}}. \qquad (8)$$

for $1 \leq i < j \leq n$ and let $\gamma_{ii} = 1$ ($1 \leq i \leq n$).

Matrix $\Gamma_n$, and its operator norm $\|\Gamma_n\|$ w.r.t. the Euclidean distance, are the measures of dependence for the random sequence $X_1, X_2, \ldots, X_n$. For example if the $X_i$s are independent, $\Gamma_n = \mathbf{I}$ and $\|\Gamma_n\| = 1$. In general $\|\Gamma_n\|$, which appears in the forthcoming concentration inequalities for dependent sequences, can grow with $n$. Since the concentration bounds are homogeneous in $n/\|\Gamma_n\|^2$, a larger value $\|\Gamma_n\|^2$ means a smaller "effective" sample size. This motivates the following definition.

**Definition 4.** *We say that a time-homogeneous Markov chain uniformly quickly forgets its past if $\tau = \sup_{n \geq 1} \|\Gamma_n\|^2 < +\infty$. Further, $\tau$ is called the forgetting time of the chain.*

Conditions under which a Markov chain uniformly quickly forgets its past are of major interest. The following proposition, extracted from the discussion on pages 421–422 of the paper by Samson (2000), gives such a condition.

**Proposition 5.** *Let $\mu$ be some nonnegative measure on $\mathcal{X}$ with nonzero mass $\mu_0$. Let $P^i$ be the $i$-step transition kernel as defined above. Assume that there exists some integer $r$ such that for all $x \in \mathcal{X}$ and all measurable sets $A$,*

$$P^r(A|x) \leq \mu(A). \qquad (9)$$

*Then,*

$$\|\Gamma_n\| \leq \frac{\sqrt{2}}{1 - \rho^{\frac{1}{2r}}},$$

*where $\rho = 1 - \mu_0$.*

Meyn and Tweedie (2009) call homogeneous Markov chains that satisfy the majorization condition (9) *uniformly ergodic*. We note in passing that there are other cases when $\|\Gamma_n\|$ is known to be independent of $n$. Most notable, this holds when the Markov chain is contracting. The matrix $\Gamma_n$ can also be defined for more general dependent processes and such that the theorem below remains valid. With such a definition, $\|\Gamma_n\|$ can be shown to be bounded for general $\Phi$-dependent processes.

The following result is a trivial corollary of Theorem 2 of Samson (2000) (Theorem 2 is stated for empirical processes and can be considered as a generalization of Talagrand's inequality to dependent random variables):

**Theorem 6.** *Let $f$ be a measurable function on $\mathcal{X}$ whose values lie in $[0, B]$, $(X_i)_{1 \leq i \leq n}$ be a homogeneous Markov chain taking values in $\mathcal{X}$ and let $\Gamma_n$ be the matrix with elements defined by* (8). *Let*

$$Z = \frac{1}{n} \sum_{i=1}^n f(X_i).$$

*Then, for every $\varepsilon \geq 0$,*

$$\mathbb{P}(Z - \mathbb{E}[Z] \geq \varepsilon) \leq \exp\left(-\frac{\varepsilon^2 n}{2B \|\Gamma_n\|^2 (\mathbb{E}[Z] + \varepsilon)}\right),$$

$$\mathbb{P}(\mathbb{E}[Z] - Z \geq \varepsilon) \leq \exp\left(-\frac{\varepsilon^2 n}{2B \|\Gamma_n\|^2 \mathbb{E}[Z]}\right).$$

Lemma 2 is an immediate consequence of this theorem.